\documentclass[runningheads]{llncs}
\usepackage[T1]{fontenc}
\usepackage{graphicx,verbatim}
\usepackage{url}
\usepackage{amsmath}
\usepackage{booktabs}
\usepackage[hidelinks]{hyperref}
\usepackage{orcidlink}
\usepackage{bbding}
\renewcommand{\orcidID}[1]{\orcidlink{#1}}
\begin{document}
\title{Semantic-Guided Multimodal Preprocessing for Vision Transformer-Based Clear Cell Renal Cell Carcinoma Grading}
\titlerunning{Semantic-Guided Preprocessing for ViT-Based CCRCC Grading}

\author{Fatemeh Javadian\inst{1}\orcidID{0009-0001-6269-5237}\Envelope \and
Zhu Chen\inst{1,2}\orcidID{0009-0009-9847-7686} \and
Zahra Aminparast\inst{3}\orcidID{0000-0002-4704-8592} \and
Johannes Stegmaier\inst{1,2}\orcidID{0000-0003-4072-3759}}
\authorrunning{F. Javadian et al.}
\institute{Institute of Imaging and Computer Vision, RWTH Aachen University, Aachen, Germany \\
    \email{fatemeh.javadian@rwth-aachen.de} \and
    Machine Learning for Medical Data, Faculty of Mathematics and Natural Sciences, Heinrich Heine University Düsseldorf, Düsseldorf, Germany \and
    Institute of Medical Sciences, Kermanshah University, Medical School, Kermanshah, Iran}

\maketitle
\begin{abstract}
Clear cell renal cell carcinoma (CCRCC) grading is essential for treatment planning, yet existing approaches either analyze patch-level images directly or focus solely on nuclei-level classification, without linking to final tumor grading. We propose a semantic-guided multimodal preprocessing method that integrates nuclei classification maps from existing pre-trained models with RGB histopathology images for Vision Transformer (ViT)-based CCRCC grading. Our approach employs classification map channel concatenation and multiplicative modulation, with optimized overlays to leverage nuclei grading information, while preserving RGB textural features. Evaluation of multiple preprocessing strategies demonstrates that semantic-guided enhancement achieves 0.916 balanced accuracy, outperforming RGB-only baseline (0.707) and max-voting aggregation from prior studies (0.427). Sensitivity analysis reveals that this 21 percentage point improvement over baseline persists even under simulated perturbation at rates matching current state-of-the-art nuclei classification model error thresholds, suggesting both effective semantic utilization and practical robustness. These findings show that preprocessing-based multimodal fusion can leverage the diagnostic potential of existing imperfect nuclei classifiers, effectively bridging previously isolated fine-grained nuclear-level analysis with coarse-grained ViT-based patch classification. Per-class recall was consistent across grades (0.93, 0.91, 0.91), indicating that gains are not concentrated in the majority class. Because the sensitivity analysis perturbs ground-truth maps rather than predictions from an actual nuclei model, this result characterizes robustness under simulated error rather than deployment with a real upstream model, which remains for future work.

\end{abstract}
\section{Introduction}

Clear cell renal cell carcinoma (CCRCC) is a common subtype of renal cancer, presenting complex histological features that challenge diagnosis and grading \cite{Browning2021}. Grading guides assessment of tumor aggressiveness and treatment decisions \cite{Warren2018,Choi2024}. The World Health Organization/International Society of Urological Pathology (WHO/ISUP) system grades CCRCC by nucleolar prominence, with higher grades indicating greater aggressiveness~\cite{Delahunt2019}. Grades~1--3 form a progressive, nucleus-centered continuum defined by reproducible nuclear features at specific magnifications. Grade~4 represents extreme dedifferentiation (sarcomatoid or rhabdoid morphology) where nuclei no longer follow consistent patterns, and is therefore analyzed separately~\cite{Warren2018,Delahunt2019}. Among nucleolar-based grades, grade~3 is the most clinically significant, followed by grades~2 and~1~\cite{Warren2018,Delahunt2019}. Whole Slide Images (WSIs) are typically segmented into patches that are graded to infer overall tumor grade \cite{Campanella2019}. The WHO/ISUP system lacks formal quantitative thresholds; pathologists instead weigh the presence, spatial distribution, and morphology of high-grade nuclei \cite{Mercan2022,Paech2011}. This qualitative process shows significant inter-observer variability \cite{Browning2021,Paech2011}, limiting simple aggregation and requiring patch evaluation of both individual nuclei and tissue context \cite{Campanella2019}.

Patch-level strategies are either coarse-grained, grading entire patches directly \cite{Tabibu2019,Chanchal2023}, or fine-grained, grading individual nuclei then aggregating \cite{Gao2021}. Fine-grained classification captures tumor-cell heterogeneity and subtle morphological variation \cite{Graham2019,Zhou2018}, but aggregating nuclei-level grades into patch predictions remains difficult. Many methods use max-voting, assigning the most abundant nuclei class as the patch grade~\cite{Gao2021,Bilal2023}, which systematically under-grades patches where sparse high-grade nuclei determine clinical significance~\cite{Chen2021graph,Campanella2019}.

\begin{figure}[t]
\centering
\includegraphics[width=0.19\linewidth]{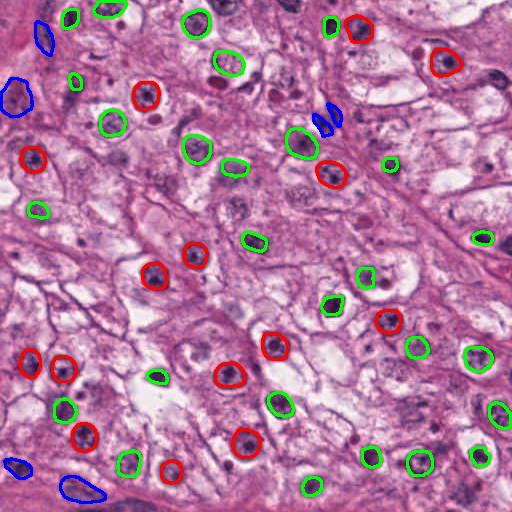}
\hfill
\includegraphics[width=0.19\linewidth]{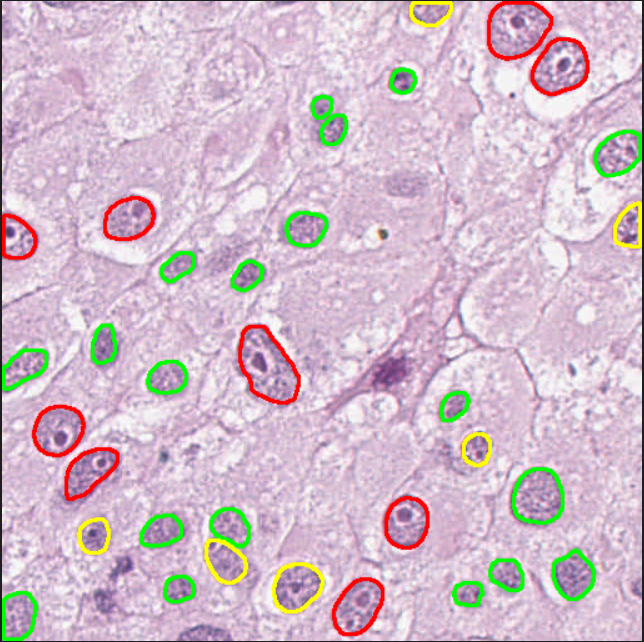}
\hfill
\includegraphics[width=0.19\linewidth]{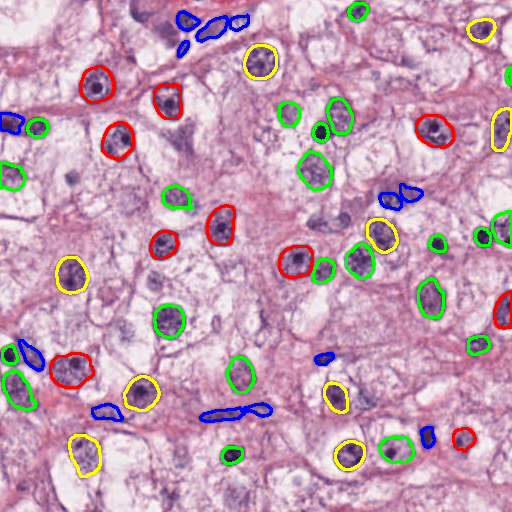}
\hfill
\includegraphics[width=0.19\linewidth]{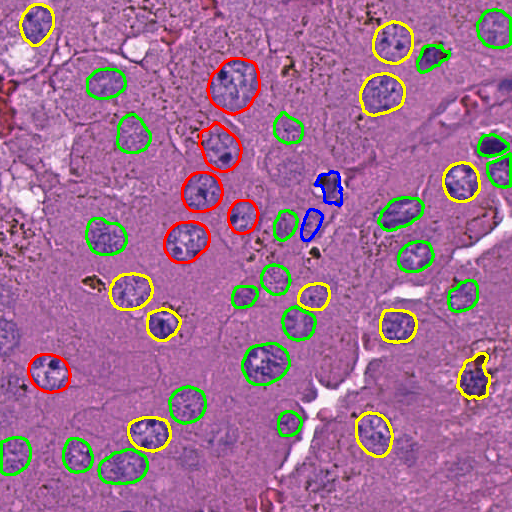}
\hfill
\includegraphics[width=0.19\linewidth]{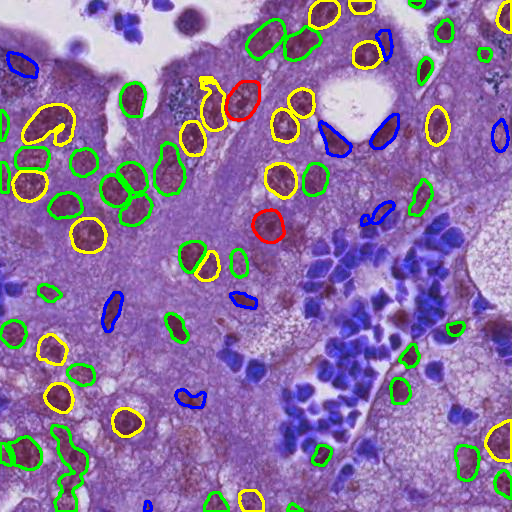}
\caption{Limitation of max-voting for CCRCC patch grading. Nuclei are outlined by 4 grades: blue (grade 0, non-tumorous cells), green (grade 1), yellow (grade 2), red (grade 3). \textbf{Images 1-4 (left):} Despite predominance of low-grade nuclei (green/yellow), all four patches are assigned grade 3 by the pathologist due to presence and spatial distribution of high-grade nuclei (red). Max-voting (selecting the most abundant grade) would incorrectly classify these as grade 1 or 2. \textbf{Image 5 (right):} A grade 2 patch showing that mere presence of scattered grade 3 nuclei is insufficient, grade assignment depends on quantity, spatial distribution, and morphological context, determined subjectively by the pathologist.}
\label{fig:image_compare}
\end{figure}

Fig.~\ref{fig:image_compare} illustrates this limitation. Images 1--4 are grade~3 patches in which green (grade~1) and yellow (grade~2) nuclei far outnumber red (grade~3) nuclei; max-voting~\cite{Gao2021} would misclassify them as grade~1 or~2, whereas pathologists assign grade~3 from the presence and spatial distribution of high-grade nuclei even as a minority. Conversely, Image~5 is a grade~2 patch with scattered grade~3 nuclei, showing that mere presence of high-grade nuclei is insufficient: spatial clustering, density, and morphological context also matter. These subjective, non-linear relationships cannot be captured by simple voting.

\subsection{State of the Art}
\label{ssec:art}
U-Net architectures and variants~\cite{Rasheed2023,Senapati2024,Castro2024,Graham2019} yield accurate per-nucleus grade predictions, but aggregating them via max-voting discards the spatial and morphological context essential for clinical grading~\cite{Bilal2023}. Separately, coarse-grained methods grade patches directly using deep learning, particularly Vision Transformers (ViTs) that capture local and global patterns~\cite{Chen2022hipt,Shao2021transmil}, bypassing nuclei-level analysis and relying on RGB images alone. A critical gap remains: fine-grained nuclei classifiers and coarse-grained patch classifiers stay isolated. Pre-trained nuclei models~\cite{Gao2021} encode substantial cellular-morphology knowledge unused by patch-level systems. Multimodal fusion in computational pathology~\cite{Chen2022pathomic} integrates complementary sources, but preprocessing-based fusion of nuclei maps with RGB images for patch grading remains, to our knowledge, unexplored.

We bridge this gap with a semantic-guided multimodal preprocessing method\footnote{We use \emph{multimodal} to denote the fusion of the RGB image with a derived semantic representation of the same tissue section, rather than distinct imaging modalities; both inputs originate from the same H\&E image.} integrating nuclei classification maps from pre-trained models with RGB images for ViT-based CCRCC grading. Multiplicative modulation with perceptually optimized overlays embeds nuclei grading information directly into ViT inputs, letting the model learn grade-determining patterns that combine fine-grained nuclei information with tissue context. Sensitivity analyses quantify how imperfect nuclei classification affects patch-level grading, indicating applicability with existing models. Dataset details are in Section~\ref{ssec:dataset}; subsequent sections present the methodology, results, and workflow implications.
\section{Methodology}
\subsection{Dataset}
\label{ssec:dataset}
The dataset comprises 1000 H\&E stained patches at $512 \times 512$ pixels, with patch-level ground truth annotated by a pathologist following WHO/ISUP guidelines~\cite{Browning2021,Warren2018} for grades~1--3, enabling comparison with aggregation methods such as max-voting. Single-pathologist annotation disregards inter-observer variability; our focus is the methodological advantage of semantic-guided preprocessing, not clinical-grade ground truth. The original distribution was highly imbalanced (grade~1: 66.3\%, grade~2: 23.0\%, grade~3: 10.7\%). Data were split into training (70\%), validation (10\%), and test (20\%). The patches were selected from whole-slide images of the TCGA KIRC and KIRP projects, following the patch selection and annotation of~\cite{Gao2021}. The source dataset is distributed as a flat collection of patches without an explicit patch-to-slide mapping, and its own description specifies only that the data were randomly divided into training, validation and test partitions, without stating whether this division was performed at the slide, case or patch level. Our split therefore follows the same patch-level protocol to remain comparable with the source benchmark, and slide-level grouping could not be verified or enforced. Consequently, patches originating from the same slide may appear in different partitions, and residual correlation arising from shared staining, scanner characteristics and local tissue architecture cannot be excluded. To address imbalance, targeted augmentation (horizontal/vertical flips) was applied to minority classes in the training set only, yielding a near-balanced training distribution (grade~1: 35.7\%, grade~2: 35.1\%, grade~3: 29.2\%).

\begin{table}[htbp]
\centering
\caption{Vision Transformer model comparison for patch-level CCRCC classification on RGB-only histopathology data (without semantic guidance). All models were fine-tuned with identical protocol; performance shown on test set.}
\label{tab:vit_comparison}
\setlength{\tabcolsep}{4pt}
\small
\begin{tabular}{|l|c|c|c|c|}
\hline
\textbf{Model} & \textbf{F1} & \textbf{Accuracy} & \textbf{Precision} & \textbf{Recall} \\
\hline
Google ViT Base Patch32-384 & \textbf{0.8095} & \textbf{0.8077} & \textbf{0.8181} & \textbf{0.8077} \\
Google ViT Large Patch32-384 & 0.8079 & 0.8076 & 0.8112 & 0.8076 \\
Google ViT Base Patch16-224 & 0.7766 & 0.7821 & 0.7846 & 0.7821 \\
Google ViT Base Patch16-224-in21k & 0.7536 & 0.7564 & 0.7546 & 0.7564 \\
OpenAI CLIP ViT Base Patch32 & 0.5454 & 0.5513 & 0.5538 & 0.5513 \\
\hline
\end{tabular}
\end{table}

\subsection{Vision Transformer Selection and Fine-Tuning}
\label{ssec:vit_selection}

We fine-tuned multiple pre-trained ViT architectures on our CCRCC dataset using RGB images only (without semantic guidance), replacing the classification head with a three-class output layer and updating all weights. All patches were resized to $384 \times 384$ pixels. Table~\ref{tab:vit_comparison} compares performance. Google ViT Base Patch32-384, pre-trained on ImageNet-21k, achieved the highest performance and was selected for all experiments; it processes images as $32 \times 32$ pixel-token sequences, capturing local and global tissue context~\cite{Chen2022hipt,Shao2021transmil}. Fine-tuning used learning rate $10^{-4}$ with cosine annealing, batch size 32, AdamW, and 50 epochs with early stopping. The same protocol was applied to all multimodal fusion methods below.

\subsection{Semantic-Guided Multimodal Preprocessing}
\label{ssec:preprocessing}

Our approach integrates nuclei classification maps with RGB images through preprocessing, letting the ViT leverage semantic nuclei information without architectural modifications. Maps from pre-trained models~\cite{Javadian2025,Gao2021} are combined with RGB patches before input to the fine-tuned ViT. We developed two methods: Classification Map Channel Concatenation and Multiplicative Modulation.

\subsubsection{Classification Map Channel Concatenation}

The classification map channel concatenation (HEC) method uses color deconvolution~\cite{Ruifrok2001} to separate H\&E staining components, creating a three-channel input where nuclei grading information replaces the original blue channel. RGB patches are converted to hematoxylin (H) and eosin (E) grayscale channels via standard deconvolution matrices~\cite{Ruifrok2001}. The third channel (C) holds the nuclei classification map with five values: background (0), tumor grades 1--3, and grade~0 non-tumorous cells (4). Value~4 denotes a non-tumorous cell label, unrelated to WHO/ISUP grade~4, which is excluded here. Values are linearly mapped to $[0,255]$ ($\text{output} = \text{class} \times 63.75$) for compatibility with standard image processing. The ViT is fine-tuned on HEC inputs using the RGB-only protocol.

\subsubsection{Multiplicative Modulation Method}

\begin{figure}[htb]
\centering
\includegraphics[width=0.19\linewidth]{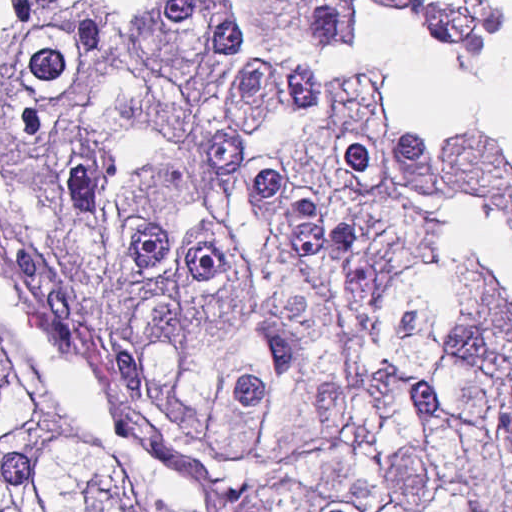}\hfill
\includegraphics[width=0.19\linewidth]{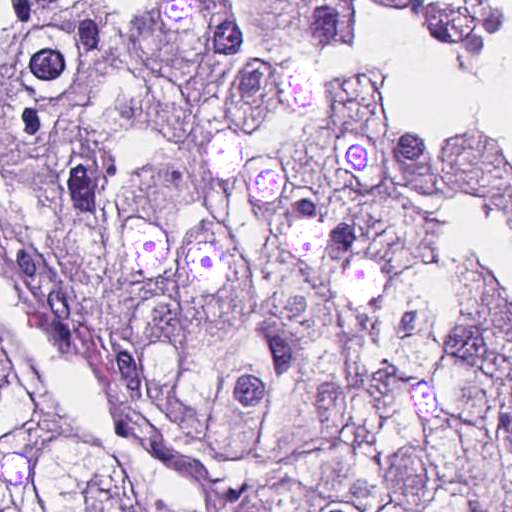}\hfill
\includegraphics[width=0.19\linewidth]{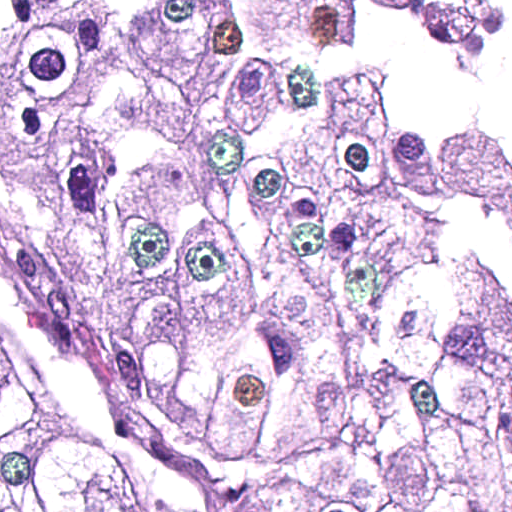}\hfill
\includegraphics[width=0.19\linewidth]{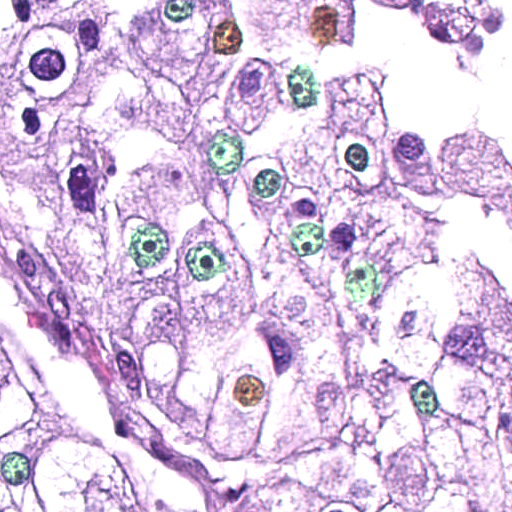}\hfill
\includegraphics[width=0.19\linewidth]{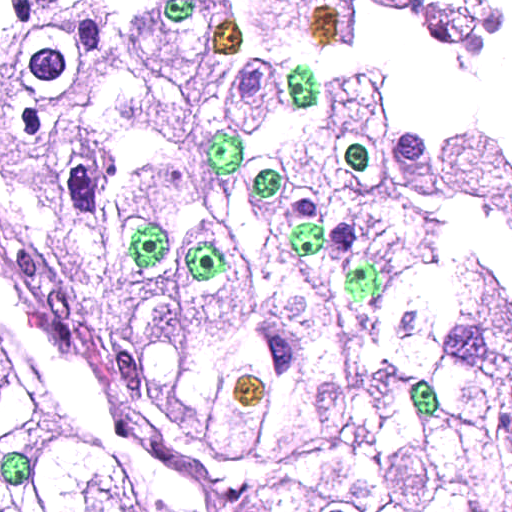}
\noindent
\makebox[0.19\linewidth]{(a)}\hfill
\makebox[0.19\linewidth]{(b)}\hfill
\makebox[0.19\linewidth]{(c)}\hfill
\makebox[0.19\linewidth]{(d)}\hfill
\makebox[0.19\linewidth]{(e)}
\caption{Preprocessing examples from left to right: (a) Original RGB image, (b) MM ($\alpha=0.9$, $\beta=5.0$, $\sigma=1.5$, overlay=0), (c) MM ($\alpha=0.5$, $\beta=2.0$, $\sigma=2.0$, overlay=0.2), (d) MM ($\alpha=0.7$, $\beta=2.5$, $\sigma=2.0$, overlay=0.35), (e) MM ($\alpha=0.85$, $\beta=3.0$, $\sigma=1.5$, overlay=0.5).}
\label{fig:five_plots}
\end{figure}

HEC combines modalities but requires color deconvolution and does not explicitly emphasize clinically important grades. Multiplicative Modulation (MM) is a single configurable method that preserves the original RGB image while emphasizing tumor regions by nuclei grade importance, grounded in information-theoretic fusion principles~\cite{Chen2022pathomic}. Building on multiplicative enhancement in medical imaging~\cite{Mou2024,Mahapatra2020}, particularly Retinex-based reflectance/illumination decomposition, MM unifies four components (intensity modulation, grade-specific weighting, spatial smoothing, and color overlay) into one preprocessing step applied before ViT input (Fig.~\ref{fig:five_plots}).

\noindent \textbf{Intensity Modulation:} For each pixel $(x,y)$, enhanced intensity is
\begin{equation}
I'(x,y) = I(x,y) \cdot (1 + \alpha \cdot f(C(x,y)))
\label{eq:mult_mod2}
\end{equation}
where $I(x,y)$ is the original RGB intensity, $C(x,y) \in \{0,1,2,3,4\}$ is the nuclei classification label (0=background, 1--3=tumor grades, 4=non-tumorous $\equiv$ grade~0), $f$ is the grade-dependent weighting below, and $\alpha > 0$ controls modulation strength. Higher $\alpha$ increases nuclei influence; as $\alpha \to 0$, the output approaches the original RGB. The multiplicative form preserves original gradients, $\nabla I' = (1 + \alpha \cdot w_c)\nabla I$, scaling rather than eliminating edge information and maintaining texture features critical for ViT attention~\cite{He2010} while adding semantic emphasis via the classification map~\cite{Xu2011}.

\noindent \textbf{Grade-Dependent Weighting:} The weighting $f$ in Eq.~\ref{eq:sigmoid} uses a sigmoid for smooth grade transitions, following ordinal classification principles for hierarchical medical grading~\cite{LeVuong2021,Zhang2023neuro}:
\begin{equation}
f(c) = \frac{\exp(\beta \cdot (c - c_0))}{1 + \exp(\beta \cdot (c - c_0))}
\label{eq:sigmoid}
\end{equation}
where $\beta$ controls steepness and $c_0 = 1.5$ positions the midpoint between grades~1 and~2 so grade~3 receives maximum emphasis. The sigmoid applies only to tumor grades (1--3), mapping each to a weight in $[0,1]$ with $f(3) \approx 1$, reflecting grade~3 as most clinically significant~\cite{Warren2018,Delahunt2019}. Background and non-tumorous cells receive fixed weights outside the sigmoid. In practice, the sigmoid combined with grade-specific base weights $w_c$ yields a clinical-importance hierarchy: background/non-tumorous $w_0 = w_4 = 1.00$; grade~1 $w_1 = 1.10$--$1.25$; grade~2 $w_2 = 1.40$--$1.80$; grade~3 $w_3 = 1.85$--$2.0$. These are applied in Eq.~\ref{eq:mult_mod2} through $f(C(x,y))$.

\noindent \textbf{Spatial Smoothing:} To avoid harsh boundaries between adjacent nuclei regions, Gaussian smoothing ($\sigma = 1.5$--$2.0$ pixels)~\cite{Orlando2017} is applied to the classification map before fusion, operating only on the classification channel, not the RGB data, preserving tissue texture while softening region boundaries.

\noindent \textbf{Color Overlay:} Optionally, perceptually optimized color overlays~\cite{Li2015} are blended with the modulated image using coefficient $O \in [0,1]$, output $(1-O)\cdot I + O \cdot C_{\text{color}}\cdot I$, practical range 0.2--0.5. Distinct colors mark each grade: green (1), yellow (2), red (3) (Fig.~\ref{fig:five_plots}), giving the model explicit class distinction beyond intensity differences. Fig.~\ref{fig:five_plots}(b) shows a no-overlay case: modulation and weighting still enhance, but grades remain visually similar, giving spatial segmentation without color-based distinction. Overlay-free configurations achieve notably lower balanced accuracy (Table~\ref{tab:test_performance}), indicating color overlay is the primary carrier of class-discriminative information.

\noindent \textbf{Parameter Selection:} We evaluated MM configurations by grid search over $\alpha$ (0.5--0.9), $\beta$ (2.0--5.0), $\sigma$ (0--2.0), overlay (0--0.5), and grade weights, assessing first on validation; Table~\ref{tab:test_performance} reports test results for top configurations. The best ($\alpha=0.85$, $\beta=3.0$, $\sigma=1.5$, overlay=0.5) reached 0.9160 balanced accuracy versus 0.707 for the identically trained RGB-only baseline.

\begin{table*}[t]
\centering
\caption{Test Set Performance of Preprocessing Methods (sorted by Balanced Accuracy). MM: Multiplicative Modulation with parameters $\alpha$ (modulation strength), $\beta$ (sigmoid steepness), $\sigma$ (spatial smoothing), O (color overlay). Precision, Recall, and Accuracy are micro-averaged; F1 is macro-averaged. All MM configurations maintain high recall, ensuring critical high-grade cases are not missed. \textbf{Bold} indicates best performance per metric; \underline{underline} indicates second best.}
\label{tab:test_performance}
\begin{tabular}{|l|c|c|c|c|c|c|c|}
\hline
\textbf{Model} & \textbf{$\alpha$} & \textbf{$\beta$} & \textbf{Bal. Acc.} & \textbf{Accuracy} & \textbf{Precision} & \textbf{Recall} & \textbf{F1} \\
\hline
Original RGB & N/A & N/A & 0.7071 & 0.7723 & 0.7723 & 0.7723 & 0.7611 \\
HEC & N/A & N/A & 0.8612 & 0.8911 & 0.8911 & 0.8911 & 0.8859 \\
\hline
MM ($\sigma$=1.5, O=0.5) & 0.85 & 3.0 & \textbf{0.9160} & \underline{0.9208} & \underline{0.9208} & \underline{0.9208} & \underline{0.9220} \\
MM ($\sigma$=2.0, O=0.3) & 0.65 & 2.5 & \underline{0.9125} & \textbf{0.9307} & \textbf{0.9307} & \textbf{0.9307} & \textbf{0.9301} \\
MM ($\sigma$=2.0, O=0.35) & 0.70 & 2.5 & 0.9067 & 0.9208 & 0.9208 & 0.9208 & 0.9219 \\
MM ($\sigma$=0.0, O=0.4) & 0.70 & 3.0 & 0.9009 & 0.9109 & 0.9109 & 0.9109 & 0.9129 \\
MM ($\sigma$=2.0, O=0.4) & 0.75 & 3.0 & 0.9009 & 0.9109 & 0.9109 & 0.9109 & 0.9129 \\
MM ($\sigma$=2.0, O=0.2) & 0.50 & 2.0 & 0.8822 & 0.9109 & 0.9109 & 0.9109 & 0.9103 \\
MM ($\sigma$=2.0, O=0) & 0.80 & 4.0 & 0.7889 & 0.8317 & 0.8317 & 0.8317 & 0.8240 \\
MM ($\sigma$=1.5, O=0) & 0.90 & 5.0 & 0.7703 & 0.8317 & 0.8317 & 0.8317 & 0.8195 \\
\hline
\end{tabular}
\end{table*}

\subsection{Sensitivity Analysis}
\label{ssec:sensitivity}
Real-world deployment relies on pre-trained nuclei model predictions~\cite{Gao2021,Javadian2025} rather than ground truth. We tested robustness to imperfect classification by injecting controlled errors into classification maps: (1) segmentation errors: random nuclei nullified, simulating false negatives; (2) classification errors: grades randomly altered with adjacent-grade preference (e.g., grade~2 $\leftrightarrow$ grade~3), reflecting realistic misclassification~\cite{Javadian2025}. Both were applied simultaneously from 0\% to 60\%. Models were trained on ground truth; perturbation applied only at evaluation, isolating nuclei-accuracy impact while holding RGB constant. Performance used balanced accuracy and F1.

This design was chosen deliberately: applying controlled perturbations to ground-truth maps yields a tunable error rate and therefore a continuous characterization of how performance depends on map accuracy, whereas the predictions of a single fixed model provide only one operating point and confound segmentation with classification errors. The perturbation range was calibrated against the error rates measured when running the pre-trained models~\cite{Gao2021,Javadian2025} on this dataset, so that the tested interval brackets the accuracy of current nuclei classifiers. Two limitations follow. First, randomly injected errors do not reproduce the structured error patterns of a real upstream model, whose mistakes are correlated with difficult morphology, staining variation, nuclear density and grade, and which may therefore concentrate on precisely the high-grade nuclei that determine the patch label. Second, models were trained on ground-truth maps and perturbed only at evaluation, so the reported degradation characterizes robustness to corruption of a reliable input rather than performance with an imperfect model in the loop at both training and evaluation time. Evaluating the full pipeline with predicted maps in both phases is left for future work. Balanced accuracy is average per-class recall:
\begin{equation}
\text{Balanced Accuracy} = \frac{1}{N} \sum_{i=1}^{N} \frac{\text{TP}_i}{\text{TP}_i + \text{FN}_i}
\end{equation}
where $N$ is the number of classes, giving equal importance to all grades regardless of frequency and addressing dataset imbalance~\cite{Brodersen2010}.

\begin{figure}[htb]
\centering
\includegraphics[width=\linewidth]{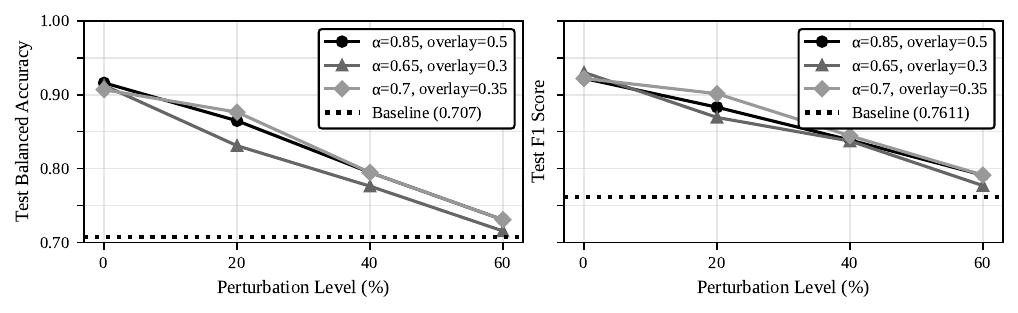}
\caption{Perturbation impact on balanced accuracy and F1 score. Performance degrades with increasing simultaneous segmentation and classification errors but remains above RGB-only baselines (0.707 and 0.761, respectively) up to 60\% error rate, demonstrating robustness to imperfect nuclei predictions.}
\label{fig:perturbation}
\end{figure}

\section{Results and Discussion}
\label{sec:results}
We evaluated two fusion methods, HEC and MM. Table~\ref{tab:test_performance} shows all optimized methods outperform the RGB-only baseline, with MM superior by preserving gradient information while directly modulating RGB intensities without color deconvolution. The top MM configurations confirm the ranges in Section~\ref{ssec:preprocessing}: $\alpha = 0.7$--$0.85$ was best (lower under-emphasized nuclei, higher over-saturated), $\beta = 2.5$--$3.0$ balanced smoothness and separability, $w_3 = 1.85$--$2.0$ strengthened the key class without degrading texture, $\sigma = 1.5$--$2.0$ softened boundaries (minimal at $\sigma=0$). Color overlay strength (0.3--0.5) was the most influential parameter, producing perceptually distinct markers against the H\&E background; removing it significantly reduced balanced accuracy (Table~\ref{tab:test_performance}), confirming it as the primary carrier of explicit class-discriminative information, with only spatial cues remaining otherwise (Fig.~\ref{fig:five_plots}(b)).

Our best method reached 0.916 balanced accuracy and 0.922 F1, versus the RGB-only baseline (0.707, 0.761) and max-voting aggregation (0.427)~\cite{Gao2021}, confirming that max-voting under-grades patches where sparse high-grade nuclei carry clinical significance (Fig.~\ref{fig:image_compare}). 

\begin{table}[htbp]
\centering
\caption{Per-class recall of the best configuration (MM, $\alpha=0.85$) on the test set.}
\label{tab:per_class_recall}
\setlength{\tabcolsep}{4pt}
\small
\begin{tabular}{|l|c|c|c|}
\hline
\textbf{Grade} & \textbf{Test patches ($n$)} & \textbf{Correctly classified} & \textbf{Recall} \\
\hline
1 & 114 & 106 & 0.930 \\
2 & 55 & 50 & 0.909 \\
3 & 33 & 30 & 0.909 \\
\hline
Total & 202 & 186 & 0.921 \\
\hline
\end{tabular}
\end{table}

Table~\ref{tab:per_class_recall} reports per-class recall for this configuration, showing that performance is not concentrated in the majority class. Classification maps alone (with ground truth) also reach 0.916, but assume perfect classification unavailable in practice. Current nuclei models achieve $\approx$0.64--0.70 balanced accuracy~\cite{Gao2021,Javadian2025}, i.e., 30--36\% error; at 30\% perturbation our fusion maintains 0.82--0.86, indicating RGB texture provides complementary robustness.

Fig.~\ref{fig:perturbation} reports sensitivity results with segmentation and classification errors applied simultaneously. Even at 60\% combined perturbation, performance stayed above the RGB-only baseline, demonstrating applicability with existing imperfect models. When semantic modulation is too weak (e.g., $\alpha=0.3$ with minimal weights), it distorts nuclei RGB appearance without sufficient emphasis and drops below baseline (0.680 vs 0.707), supporting active utilization of nuclei grading rather than general spatial attention. Preliminary experiments examined segmentation and classification errors in selected models~\cite{Gao2021,Javadian2025} in isolation. Both types produced similar degradation, with segmentation errors and lower-grade misclassification slightly less impactful than higher-grade misclassification; combined perturbation falls between the isolated cases, making it a practical summary metric, so detailed isolated per-class/per-error analysis is omitted. We attribute this robustness to the ViT leveraging nuclei-derived features alongside RGB texture: the 21 percentage point gain over baseline and monotonic degradation with perturbation both support active semantic utilization. Removing color overlay (overlay=0) eliminates explicit class distinction while preserving segmentation-level spatial information, reducing balanced accuracy to $\approx$0.78 (Table~\ref{tab:test_performance}). Removing nuclei entirely, leaving only background tissue, eliminates chromatin texture, staining intensity, and nuclear morphology while preserving only size and contour; performance drops substantially below baseline, since WHO/ISUP criteria depend on nuclear characteristics beyond size. Our fusion integrates previously isolated fine-grained classifiers with patch-level grading without architectural modification, enabling practical deployment.

\section{Conclusion}
\label{sec:conclusion}
This study assessed whether existing imperfect nuclei classification models can improve patch-level grading, and up to which error rate integration remains practically acceptable. Semantic-guided multimodal preprocessing enhances patch-level CCRCC grading without architectural modification. Integrating nuclei maps from pre-trained models with RGB images via multiplicative modulation and perceptually optimized overlays achieved substantial gains over RGB-only processing (0.707) and max-voting aggregation (0.427), the best configuration reaching 0.916 balanced accuracy (Table~\ref{tab:test_performance}). This lets ViTs exploit both fine-grained nuclei and coarse-grained tissue patterns, evidenced by consistent advantage across all configurations. Sensitivity analysis showed robustness up to 60\% combined segmentation and classification perturbation while staying above the RGB-only baseline; current state-of-the-art nuclei models~\cite{Gao2021,Javadian2025} exhibit $\approx$30--36\% error, well within this tolerance. Without architectural modification, our approach integrates previously isolated fine-grained classifiers into practical diagnostic pipelines, with further gains possible via architectural adaptation. Because the method requires only a nuclei classification map at the input stage, it can in principle be extended to new, unannotated cohorts using maps produced by an existing nuclei model, with a small annotated subset reviewed by a pathologist to verify map quality and calibrate the expected error rate. Future work will evaluate the full pipeline with predicted maps at both training and evaluation time, and will test grade-decoupled overlay variants to separate the contribution of the semantic map from that of the colour encoding.

\subsubsection{Code Availability.}
The implementation is publicly available at \href{https://github.com/javadianf/classification-semantic-fusion-vit-HP}{our GitHub repository}.

\begin{credits}
\subsubsection{\ackname}
The results published here are in whole or in part based upon data generated by the TCGA Research Network: \url{https://www.cancer.gov/tcga}.

\subsubsection{\discintname}
The authors have no competing interests to declare that are relevant to the content of this article.
\end{credits}

\section{Compliance with Ethical Standards}
\label{sec:com}
This study utilized publicly available data from the TCGA Research Network with nuclei annotations from Gao et al.~\cite{Gao2021}. No additional ethics approval was required.

%
\bibliographystyle{splncs04}
\bibliography{bib}

@article{Bilal2023,
  author = {Bilal, M. and Jewsbury, R. and Wang, R. and AlGhamdi, H. M. and Asif, A. and Eastwood, M. and Rajpoot, N.},
  title = {An aggregation of aggregation methods in computational pathology},
  journal = {Med. Image Anal.}, year = {2023}, volume = {88}, pages = {102885}}

@inproceedings{Brodersen2010,
  author = {Brodersen, K. H. and Ong, C. S. and Stephan, K. E. and Buhmann, J. M.},
  title = {The Balanced Accuracy and Its Posterior Distribution},
  booktitle = {ICPR}, year = {2010}, pages = {3121--3124}}

@article{Browning2021,
  author = {Browning, L. and Colling, R. and Verrill, C.},
  title = {{WHO/ISUP} grading of clear cell renal cell carcinoma and papillary renal cell carcinoma; validation of grading on the digital pathology platform and perspectives on reproducibility of grade},
  journal = {Diagn. Pathol.}, year = {2021}, volume = {16}, pages = {75}}

@article{Campanella2019,
  author = {Campanella, G. and Hanna, M. G. and Geneslaw, L. and Miraflor, A. and Werneck Krauss Silva, V. and Busam, K. J. and Brogi, E. and Reuter, V. E. and Klimstra, D. S. and Fuchs, T. J.},
  title = {Clinical-grade computational pathology using weakly supervised deep learning on whole slide images},
  journal = {Nat. Med.}, year = {2019}, volume = {25}, number = {8}, pages = {1301--1309}}

@article{Castro2024,
  author = {Castro, S. and Pereira, V. and Silva, R.},
  title = {Improved Segmentation of Cellular Nuclei Using {UNet} Architectures for Enhanced Pathology Imaging},
  journal = {Electronics}, year = {2024}, volume = {13}, number = {16}, pages = {3335}}

@article{Chanchal2023,
  author = {Chanchal, A. K. and Lal, S. and Kumar, R. and Kwak, J. T. and Kini, J.},
  title = {A novel dataset and efficient deep learning framework for automated grading of renal cell carcinoma from kidney histopathology images},
  journal = {Sci. Rep.}, year = {2023}, volume = {13}, pages = {5728}}

@inproceedings{Chen2021graph,
  author = {Chen, R. J. and Lu, M. Y. and Shaban, M. and Chen, C. and Chen, T. Y. and Williamson, D. F. K. and Mahmood, F.},
  title = {Whole slide images are {2D} point clouds: Context-aware survival prediction using patch-based graph convolutional networks},
  booktitle = {MICCAI}, year = {2021}, volume = {12908}, pages = {339--349}}

@inproceedings{Chen2022hipt,
  author = {Chen, R. J. and Chen, C. and Li, Y. and Chen, T. Y. and Trister, A. D. and Krishnan, R. G. and Mahmood, F.},
  title = {Scaling Vision Transformers to Gigapixel Images via Hierarchical Self-Supervised Learning},
  booktitle = {CVPR}, year = {2022}, pages = {16144--16155}}

@article{Chen2022pathomic,
  author = {Chen, R. J. and Lu, M. Y. and Wang, J. and Williamson, D. F. K. and Rodig, S. J. and Lindeman, N. I. and Mahmood, F.},
  title = {Pathomic Fusion: An Integrated Framework for Fusing Histopathology and Genomic Features for Cancer Diagnosis and Prognosis},
  journal = {IEEE Trans. Med. Imaging}, year = {2022}, volume = {41}, number = {4}, pages = {757--770}}

@article{Choi2024,
  author = {Choi, J. and Bang, S. and Suh, J. and Choi, C. I. and Song, W. and Yuk, H. D. and others},
  title = {Survival pattern of metastatic renal cell carcinoma patients according to {WHO/ISUP} grade: a long-term multi-institutional study},
  journal = {Sci. Rep.}, year = {2024}, volume = {14}, pages = {4740}}

@article{Delahunt2019,
  author = {Delahunt, B. and Eble, J. N. and Egevad, L. and Samaratunga, H.},
  title = {Grading of renal cell carcinoma},
  journal = {Histopathology}, year = {2019}, volume = {74}, number = {1}, pages = {4--17}}

@inproceedings{Gao2021,
  author = {Gao, Z. and Shi, J. and Zhang, X. and Li, Y. and Zhang, H. and Wu, J. and Wang, C. and Meng, D. and Li, C.},
  title = {Nuclei Grading of Clear Cell Renal Cell Carcinoma in Histopathological Image by Composite High-Resolution Network},
  booktitle = {MICCAI}, year = {2021}, volume = {12908}, pages = {132--142}}

@article{Graham2019,
  author = {Graham, S. and Vu, Q. D. and Raza, S. E. A. and Azam, A. and Tsang, Y. W. and Kwak, J. T. and Rajpoot, N.},
  title = {{HoVer-Net}: Simultaneous Segmentation and Classification of Nuclei in Multi-Tissue Histology Images},
  journal = {Med. Image Anal.}, year = {2019}, volume = {58}, pages = {101563}}

@article{He2010,
  author = {He, K. and Sun, J. and Tang, X.},
  title = {Guided Image Filtering},
  journal = {IEEE Trans. Pattern Anal. Mach. Intell.}, year = {2013}, volume = {35}, number = {6}, pages = {1397--1409}}

@inproceedings{Javadian2025,
  author = {Javadian, F. and Aminparast, Z. and Stegmaier, J. and Jose, A.},
  title = {Comparative Analysis of Unsupervised and Supervised Autoencoders for Nuclei Classification in Clear Cell Renal Cell Carcinoma Images},
  booktitle = {IEEE ISBI}, year = {2025}, pages = {1--5}}

@article{LeVuong2021,
  author = {Le Vuong, T. T. and Kim, K. and Song, B. and Kwak, J. T.},
  title = {Joint categorical and ordinal learning for cancer grading in pathology images},
  journal = {Med. Image Anal.}, year = {2021}, volume = {73}, pages = {102206}}

@article{Li2015,
  author = {Li, X. and Plataniotis, K. N.},
  title = {A Complete Color Normalization Approach to Histopathology Images Using Color Cues Computed From Saturation-Weighted Statistics},
  journal = {IEEE Trans. Biomed. Eng.}, year = {2015}, volume = {62}, number = {7}, pages = {1862--1873}}

@inproceedings{Mahapatra2020,
  author = {Mahapatra, D. and Bozorgtabar, B. and Thiran, J.-P. and Shao, L.},
  title = {Structure Preserving Stain Normalization of Histopathology Images Using Self Supervised Semantic Guidance},
  booktitle = {MICCAI}, year = {2020}, volume = {12265}, pages = {309--319}}

@article{Mercan2022,
  author = {Mercan, C. and Balkenhol, M. and Salgado, R. and Sherman, M. and Vielh, P. and Vreuls, W. and Pol{\'o}nia, A. and Horlings, H. M. and Weichert, W. and Carter, J. M. and Bult, P. and Christgen, M. and Denkert, C. and van de Vijver, K. and Bokhorst, J.-M. and van der Laak, J. and Ciompi, F.},
  title = {Deep learning for fully-automated nuclear pleomorphism scoring in breast cancer},
  journal = {npj Breast Cancer}, year = {2022}, volume = {8}, number = {1}, pages = {120}}

@article{Mou2024,
  author = {Mou, E. and Wang, H. and Chen, X. and Li, Z. and Cao, E. and Chen, Y. and Huang, Z. and Pang, Y.},
  title = {Retinex theory-based nonlinear luminance enhancement and denoising for low-light endoscopic images},
  journal = {BMC Med. Imaging}, year = {2024}, volume = {24}, number = {1}, pages = {207}}

@article{Orlando2017,
  author = {Orlando, J. I. and Prokofyeva, E. and Blaschko, M. B.},
  title = {A Discriminatively Trained Fully Connected Conditional Random Field Model for Blood Vessel Segmentation in Fundus Images},
  journal = {IEEE Trans. Biomed. Eng.}, year = {2017}, volume = {64}, number = {1}, pages = {16--27}}

@article{Paech2011,
  author = {Paech, D. C. and Weston, A. R. and Pavlakis, N. and Gill, A. and Rajan, N. and Barraclough, H. and Fitzgerald, B. and Van Kooten, M.},
  title = {A Systematic Review of the Interobserver Variability for Histology in the Differentiation between Squamous and Nonsquamous Non-small Cell Lung Cancer},
  journal = {J. Thorac. Oncol.}, year = {2011}, volume = {6}, number = {1}, pages = {55--63}}

@article{Rasheed2023,
  author = {Rasheed, A. and Shirazi, S. H. and Umar, A. I. and Shahzad, M. and Yousaf, W. and Khan, Z.},
  title = {Cervical cell's nucleus segmentation through an improved {UNet} architecture},
  journal = {PLoS One}, year = {2023}, volume = {18}, number = {10}, pages = {e0283568}}

@article{Ruifrok2001,
  author = {Ruifrok, A. C. and Johnston, D. A.},
  title = {Quantification of histochemical staining by color deconvolution},
  journal = {Anal. Quant. Cytol. Histol.}, year = {2001}, volume = {23}, number = {4}, pages = {291--299}}

@article{Senapati2024,
  author = {Senapati, P. and Basu, A. and Deb, M. and Dhal, K. G.},
  title = {Sharp dense {U-Net}: an enhanced dense {U-Net} architecture for nucleus segmentation},
  journal = {Int. J. Mach. Learn. Cybern.}, year = {2024}, volume = {15}, number = {6}, pages = {2079--2094}}

@inproceedings{Shao2021transmil,
  author = {Shao, Z. and Bian, H. and Chen, Y. and Wang, Y. and Zhang, J. and Ji, X. and Zhang, Y.},
  title = {{TransMIL}: Transformer based Correlated Multiple Instance Learning for Whole Slide Image Classification},
  booktitle = {NeurIPS}, year = {2021}, volume = {34}, pages = {2136--2147}}

@article{Tabibu2019,
  author = {Tabibu, S. and Vinod, P. K. and Jawahar, C. V.},
  title = {Pan-Renal Cell Carcinoma classification and survival prediction from histopathology images using deep learning},
  journal = {Sci. Rep.}, year = {2019}, volume = {9}, pages = {10509}}

@article{Warren2018,
  author = {Warren, A. Y. and Harrison, D.},
  title = {{WHO/ISUP} classification, grading and pathological staging of renal cell carcinoma: standards and controversies},
  journal = {World J. Urol.}, year = {2018}, volume = {36}, number = {12}, pages = {1913--1926}}

@article{Xu2011,
  author = {Xu, J. and Janowczyk, A. and Chandran, S. and Madabhushi, A.},
  title = {A high-throughput active contour scheme for segmentation of histopathological imagery},
  journal = {Med. Image Anal.}, year = {2011}, volume = {15}, number = {6}, pages = {851--862}}

@article{Zhang2023neuro,
  author = {Tang, W. and Yang, Z. and Song, Y.},
  title = {Disease-grading networks with ordinal regularization for medical imaging},
  journal = {Neurocomputing}, year = {2023}, volume = {545}, pages = {126245}}

@inproceedings{Zhou2018,
  author = {Zhou, Y. and Dou, Q. and Chen, H. and Qin, J. and Heng, P.-A.},
  title = {{SFCN-OPI}: Detection and Fine-grained Classification of Nuclei Using Sibling FCN with Objectness Prior Interaction},
  booktitle = {AAAI}, year = {2018}, pages = {2652--2659}}
\end{document}